\documentclass[letter, 10 pt, conference]{ieeeconf}

\IEEEoverridecommandlockouts

\usepackage{etextools}
\usepackage{amssymb}
\usepackage{float}
\usepackage{makeidx}
\usepackage{amsmath}
\usepackage{bbm}
\usepackage{epsfig}
\usepackage{epsf}
\usepackage{psfrag}
\usepackage{verbatim}
\usepackage{color}
\usepackage{multirow}
\usepackage{tabularx}
\usepackage{booktabs}
\usepackage[tight,footnotesize]{subfigure}
\usepackage{array}
\usepackage{soul}
\usepackage{footnote}
\usepackage{cite}
\usepackage{dblfloatfix}

\usepackage{color, colortbl}
\usepackage[colorlinks,bookmarksnumbered,citecolor=orange,urlcolor=orange]{hyperref}
\usepackage{graphicx}
\graphicspath{{Figures}}
\DeclareGraphicsExtensions{.pdf,.png}
\usepackage{bigstrut}
\usepackage[english]{babel}

\usepackage{csquotes}

\usepackage[T1]{fontenc}
\usepackage{algorithm, setspace}
\usepackage{algpseudocode}
\usepackage{url}
\usepackage{multirow}
\usepackage{float}
\usepackage{xcolor}
\usepackage{hyperref}
 \hypersetup{
     colorlinks=true,
     linkcolor=orange,
     filecolor=orange,
     citecolor=orange,      
     urlcolor=orange,
     }

\newcommand{\p}[1]{\smallskip \noindent \textbf{{#1}.}}

\newcommand{\fig}[1]{Figure~\ref{fig:#1}}

\usepackage{balance}
\begin{document}

\title{\LARGE
Wrapped Haptic Display for Communicating Physical Robot Learning
}

\author{
Antonio Alvarez Valdivia$^{1}$, Ritish Shailly$^{2}$, Naman Seth$^{2}$, Francesco Fuentes$^{1}$,\\ Dylan P. Losey$^{2}$, Laura H. Blumenschein$^{1}$
\thanks{This work is supported in part by the NSF Graduate Research Fellowship Program and by NSF Grant $\#2129201$.}
\thanks{$^{1}$School of Mechanical Engineering, Purdue University, Lafayette, IN 47901. 
        {e-mail: \texttt{\{alvar168, ffuente, lhblumen\}@purdue.edu}}}%
\thanks{$^{2}$These authors are members of the Collaborative Robotics Lab (\href{https://collab.me.vt.edu/}{Collab}), Dept. of Mechanical Engineering, Virginia Tech, Blacksburg, VA 24061.
\newline
{e-mail: \texttt{\{rshailly, naman7, losey\}@vt.edu}}}
}


\maketitle

\begin{abstract}
Physical interaction between humans and robots can help robots learn to perform complex tasks. The robot arm gains information by observing how the human kinesthetically guides it throughout the task. While prior works focus on how the robot learns, it is equally important that this learning is \textit{transparent} to the human teacher. Visual displays that show the robot's uncertainty can potentially communicate this information; however, we hypothesize that visual feedback mechanisms miss out on the physical connection between the human and robot. In this work we present a \textit{soft haptic display} that wraps around and conforms to the surface of a robot arm, adding a haptic signal at an existing point of contact without significantly affecting the interaction. We demonstrate how soft actuation creates a salient haptic signal while still allowing flexibility in device mounting. Using a psychophysics experiment, we show that users can \textit{accurately distinguish} inflation levels of the wrapped display with an average Weber fraction of 11.4\%. When we place the wrapped display around the arm of a robotic manipulator, users are able to interpret and leverage the haptic signal in sample robot learning tasks, improving identification of areas where the robot needs more training and enabling the user to provide better demonstrations. See videos of our device and user studies here: \href{https://youtu.be/tX-2Tqeb9Nw}{https://youtu.be/tX-2Tqeb9Nw}
\end{abstract}


\section{Introduction}

\begin{figure}[t]
	\begin{center}
		\includegraphics[width=1.0\columnwidth]{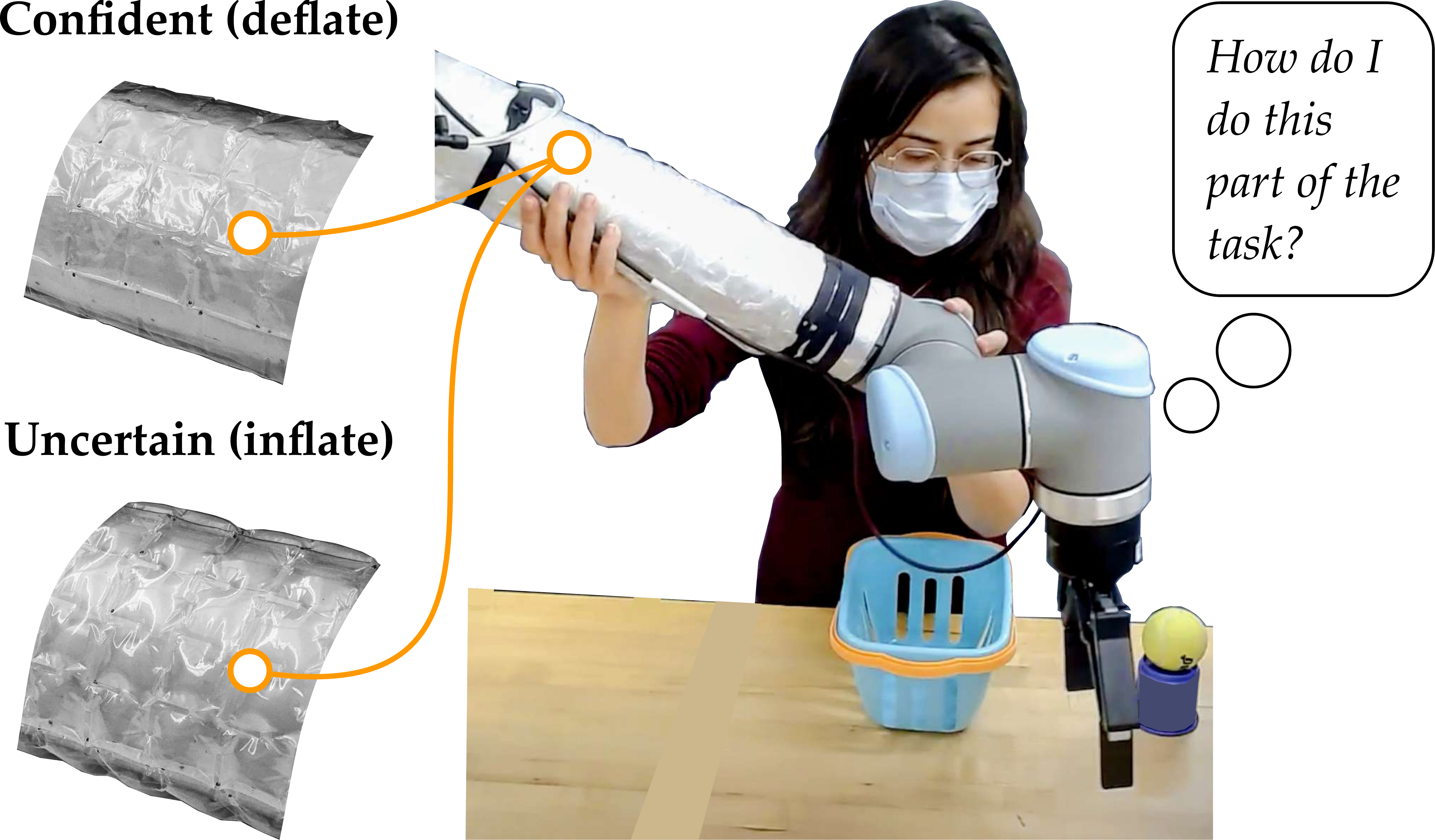}
		\vspace{-0.5em}
		\caption{Human physically teaching a robot arm. We wrap a soft pneumatic display around the arm and render haptic signals by controlling the pressure of the display. The robot learner leverages this haptic display in real-time to communicate the parts of the task that it is confident about, as well as the parts where it is uncertain and needs additional guidance.}
		\label{fig:front}
	\end{center}
 	\vspace{-2.0em}
\end{figure}

Imagine teaching a rigid robot arm to clean objects off a table (see \fig{front}). One intuitive way for you to teach this robot is through \textit{physical interaction}: you push, pull, and guide the arm along each part of the task. Of course, the robot may not learn everything from a single demonstration, and so you show multiple examples of closing shelves, removing trash, and sorting objects. As you kinesthetically teach the robot you are faced with two questions: i) has the robot learned enough to clear the table by itself and ii) if not, what parts of the task is the robot still uncertain about?

While existing work enables robots to learn from physical human interaction \cite{argall2009survey,akgun2012keyframe,pastor2009learning,losey2021physical}, having the robot effectively provide \textit{real-time feedback} to human teachers remains an open problem. Ideally, this feedback should not be cumbersome or distracting (i.e., the human must be able to focus on seamlessly guiding the robot) and should be easily interpretable (i.e., the human must be able to clearly distinguish between different signals). These requirements present a tradeoff as human fingertips provide the densest mechanoreceptors, but placing rigid devices at the hand will impact task performance. Recent research has created communication channels by wrapping \textit{haptic devices} around the human's arm \cite{che2020efficient, mullen2021communicating, dunkelberger2020multisensory}, but locating feedback on the human's body can create a disconnect with the robot's task. 

Our insight is that --- instead of asking the human teacher to wear a feedback device or watch a computer monitor ---
\begin{center}\vspace{-0.3em}
\textit{We can take advantage of the preexisting physical contact between the human and robot through slim form-factor soft haptic displays that can be \emph{wrapped} around the robot arm.}\vspace{-0.3em}
\end{center}
Accordingly, in this paper we apply soft robotics techniques to develop, analyze, and apply a wrapped haptic display for communicating robot learning. We distribute this soft display along a rigid robot arm so that wherever the human physically interacts with the robot they perceive its feedback. We then actively control the \textit{pressure} of the pneumatic display to render the robot's \textit{uncertainty}: the display inflates in regions of the task where the robot is unsure about its actions (and needs additional human teaching), and deflates in regions where the robot is confident about the task (and does not need any additional human guidance). Our hypothesis is that --- because the soft wrapped display creates a channel for communication on any surface without impacting the task --- humans will be able to more intuitively and robustly use this feedback. We experimentally demonstrate that this pressure-based feedback enables humans i) to determine whether the robot has learned enough to be deployed and ii) to identify parts of the task where kinesthetic teaching is still required.

Overall we make the following contributions:

\p{Developing Wrapped Haptic Displays} We design and build a compliant pneumatic haptic device that wraps around and conforms to the robot, providing haptic stimuli that are localized to the robot arm and distributed along its geometry. This device is manufactured using soft, flexible pouches that render haptic signals through pressure.

\p{Measuring User Perception of Wrapped Displays} We perform a psychophysics study to find the range of pressures that humans can distinguish. We report the just noticeable difference (JND) for pressures rendered by the soft display.

\p{Applying Wrapped Displays to Communicate Learning} We ask participants to kinesthetically teach a robot arm while the robot provides real-time feedback about its learning. We map the robot's uncertainty to the pressure of our wrapped display. When compared to a graphical user interface, rendering feedback with our wrapped haptic display leads to faster and more informative human teaching, and is subjectively preferred by human teachers.

\section{Related Work}

In this paper we introduce a wrapped haptic display for communicating robot learning in real-time during physical human-robot interaction. We build on previous research for kinesthetic teaching, haptic interfaces, and soft displays.

\p{Kinesthetic Teaching} Humans can show robot arms how to perform new tasks by physically demonstrating those tasks on the robot \cite{argall2009survey,akgun2012keyframe,pastor2009learning,losey2021physical}. As the human backdrives the robot, the robot records the states that it visits and the human's demonstrated actions at those states. The robot then learns to imitate the human's actions and perform the task by itself \cite{ross2011reduction}. One important outcome of the learning process is \textit{uncertainty}: the robot can measure how confidently it knows what to do at different states along the task \cite{hoque2021thriftydagger,menda2019ensembledagger}. In this paper we explore how robots should \textit{communicate} their learning uncertainty back to the human teacher. Keeping the human up-to-date with what the robot has learned builds trust and improves teaching \cite{hellstrom2018understandable}. Outside of physical human-robot interaction, prior research has developed multiple modalities to communicate robot learning and intent: these include robot motion \cite{dragan2013legibility}, graphical user interfaces \cite{huang2019enabling}, projections into the environment \cite{andersen2016projecting}, and augmented reality headsets \cite{walker2018communicating}. Within a teleoperation domain, our recent work suggests that \textit{haptic interfaces} are particularly effective at communicating low-dimensional representations of robot learning \cite{mullen2021communicating}. Here we will leverage these results to develop a real-time feedback interface \textit{specifically for} kinesthetic teaching.

\p{Haptics to Convey Intent} While haptic devices have a general goal of stimulating the human sense of touch, haptics have also been applied to communicate \textit{robot intent} or similar social features. For instance --- when studying how humans and robots should interact in shared spaces --- prior works have used haptics to explicitly convey the robot's intended direction of motion or planned actions \cite{che2020efficient,cini2021relevance,casalino2018operator}. Here the haptic feedback is worn by the human (e.g., a wristband). Wrapping the haptic device \textit{around the human's arm} enables the human to move about the space while receiving real-time feedback; but this feedback is physically separated from the task, potentially requiring additional mental energy to decode the robot's message. Recent work has shown that, given appropriate context, complex human-to-human social touch signals, like stroking \cite{nunez2020investigating,muthukumarana2020touch}, hugging \cite{HuggyPajamaTeh}, and emotional communication \cite{SalvatoTOH2021}, can be replicated and understood in a wearable format. Based on these recent successes in human-to-human touch, we revisit wrapped haptic devices: instead of locating haptic devices on the human's body, we physically constrain the haptic signal to the point of the human-robot interaction.

\p{Soft Displays} To locate the haptic signal at the point of interaction, we must develop a flexible display that can be mounted onto a wide range of robot arms without redesign. Previous soft displays have demonstrated compliance in two factors: \textit{flexibility} of the interface and \textit{compliance} of the actuators themselves. Soft haptic displays have been developed with pneumatic actuation \cite{raitor2017wrap,HapWRAP2018}, shape memory alloys \cite{muthukumarana2020touch}, and dielectric elastomers \cite{zhao2020wearable}. Many of these soft haptic displays have been physically flexible, often so that they can be worn by humans \cite{raitor2017wrap, muthukumarana2020touch}, or even wrapped around a human's arm \cite{HapWRAP2018}. We will take advantage of the intrinsic compliance of soft actuators and displays to create a wrapped haptic display that we can mount on \textit{robot arms}.

\section{Developing a Wrapped Haptic Display} \label{Haptic Display and Design}

\subsection{Requirements}\label{Requirements}
The design process for the wrapped display covered three key requirements: low volume, fast inflation, and textured surface. First, we wanted to design a display that would clearly show inflation without using large volumes of air and pressures. Limiting the volume that the display holds allows for fast inflation and deflation. This is an important design feature since fast transitions between inflation levels would allow for faster changes in the signals that the display is producing. An additional requirement was to create an inflatable surface that would produce textured tactile sensations. Our hypothesis is that a textured surface would help users to quickly identify pressure changes in the display since there are more surface features to explore with their hands. Since the target application of the wrapped haptic display is robot learning, an additional design constraint was the need to fully wrap the display around robot arms without constraining motion or impairing demonstration.

\subsection{Design}
The soft wrapped haptic display consists of an array of cells patterned into a low-density polyethylene (LDPE) plastic tube using heat sealing. The cells are interconnected to allow for smooth and fast inflation of the array via a pattern of gaps in sealing. Initial testing showed that having a single inflatable cell did not provide enough surface change to assist users in identifying pressure changes, as well as being slow to inflate. Additionally, single inflatable cells were hard to wrap around objects. Adding heat seals subdivides the cell, limiting the volume, adding additional texture, and allowing the overall surface to remain flexible when inflated and deflated. A repeated and homogeneous pattern across the entire length allowed for even and reliable inflation of the display. If the pattern was not homogeneous, we found that issues such as superfluous contraction and unintentional airflow blocking would happen.

\begin{figure}[t]
	\begin{center}
		\includegraphics[width=1\columnwidth]{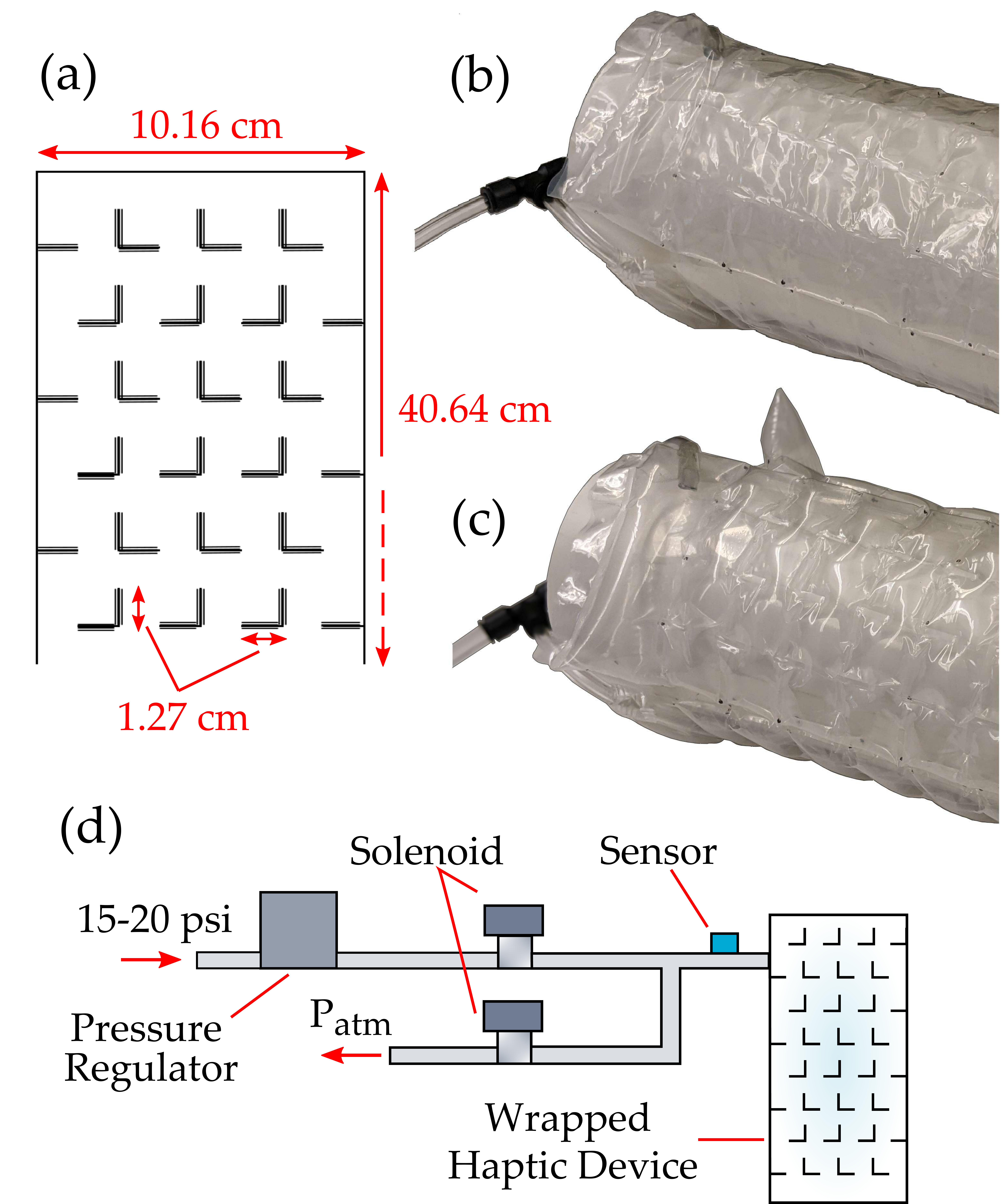}
		\caption{(a) Detailed view of the soft wrapped haptic display. The thick lines indicates places where the LDPE plastic tube was heat-sealed. The wrapped display is shown deflated (b) and inflated (c). The diagram description of the pneumatic control system is shown in (d).}
		\label{fig:device}
	\end{center}
	\vspace{-2.0em}
\end{figure}

The final square-array design is shown in Figure~\ref{fig:device}.
The soft wrapped haptic display is made from a set of 3 haptic display strips made from an LDPE plastic tube (10.16 cm wide). The plastic tube was cut to fit the length of one of the sections of the robotic arm (40.64 cm). The square pattern was manufactured into the plastic tube using a heat sealer (H-89 Foot-Operated Impulse Sealer, ULINE) that fused the plastic layers in precise lines. The sealed lines are 1.27 cm long, alternated in rows and columns to create 2.54 cm-squares during inflation. Figure~\ref{fig:device}(a) shows the design with the dimensions in more detail. Through-wall straight connectors (5779K675, McMaster-Carr) were attached to one of the sides of each strip to allow for individual inflation. Three display strips were taped together using viscoelastic adhesive tape (MD-9000, Marker Tape) to construct a sleeve that entirely wrapped the robotic arm and the pipe. 

\subsection{Implementation} \label{subsec:implementation}
The wrapped haptic display was mounted around a tube of 6.35~cm diameter, either a PVC pipe or a section of the UR10 robot arm (used for the \textbf{Wrapped} condition in Section \ref{VT User Study}). For the \textbf{Flat} condition, a strip of the haptic display was mounted on the table. The mounting arrangements fixed the wrapped display in place, restricting it to less than 10\% contraction.
Figure~\ref{fig:device}(d) shows the pneumatic control system used to implement the wrapped haptic display. A pressure regulator (QB3, Proportion-Air, McCordsville, Indiana) was controlled using an Arduino Uno via MATLAB. The Arduino sent analog signals to the QB3, which provided accurate pressure values needed for the studies. For the user study described in Section \ref{VT User Study}, the pressure regulator (550-AID, ControlAir, Amherst, New Hampshire) was controlled using the UR10's I/O controller. Airflow to the haptic display was controlled using two on-off solenoid valves (ASCO Z134A, Emerson, St. Louis, Missouri). One of these would allow airflow from the pressure regulator into the display, while the other allowed air to escape to the atmosphere for deflation. The inflation pressure was measured using an electronic pressure sensor (015PGAA5, Honeywell Sensing, Gold Valley, Minnesota).

Tests showed that the soft wrapped haptic display can be inflated quickly; pressures above 1.5 psi (10.43 kPa) inflate the display in 0.86 seconds. The display can operate to a maximum of 3.5 psi (24.13 kPa). Above that pressure the heat-sealed edges begin to tear, producing small air leaks.  
\section{Measuring Human Perception of \\ Wrapped Haptic Displays} \label{P User Study}

Understanding the human sensory perception of the soft display, especially as it compares to rigid haptic displays, is essential in determining how to apply and control the wrapped haptic display. 
To that end, we conducted a psychometric user study to measure the basic ability to distinguish touch sensations outside of the context of the target application scenario and to obtain qualitative data of how users perceive the display. Participants physically interacted with the display and were asked to distinguish between pairs of pressures. We focused on studying the user’s ability to differentiate pressure inflation levels in the display to understand the minimum pressure differential that can produce clear haptic signals. 

\subsection{Experiment Setup} \label{Exp Setup P}

\begin{figure}[t]
	\begin{center}
		\includegraphics[width=1.0\columnwidth]{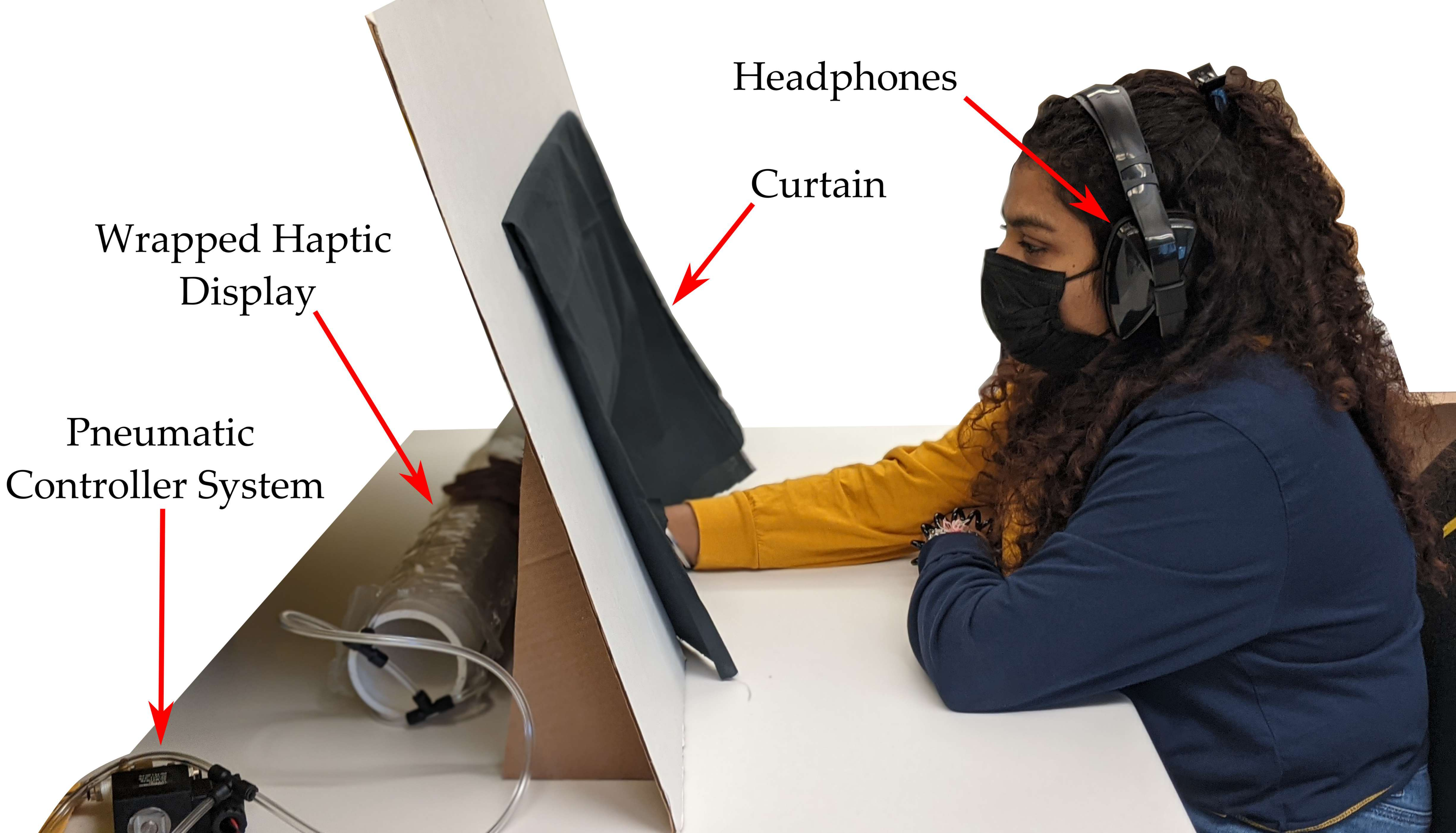}
		\caption{Experimental Setup. The participants were instructed to sit in the desk right in front of the curtain and put on hearing protection headphones.}
		\label{fig:exp_setup}
	\end{center}
	\vspace{-2.0em}
\end{figure}

The inflatable haptic display was mounted on a PVC pipe of identical diameter to the UR-10 used in Section~\ref{VT User Study}, as described in Section~\ref{subsec:implementation}. The pipe was placed lying flat and secured to a table. As shown in Figure~\ref{fig:exp_setup}, we placed a curtain to block the user’s vision and instructed users to wear hearing protection to ensure the perception study was focused entirely on tactile sensations.

The study was conducted as a forced-choice comparison where participants were asked to identify the higher pressure. The pressures were shown in pairs (i.e., reference pressure, $P_o$, vs. test pressure, $P$) to the user, distinguished as "Pressure 1" and "Pressure 2". We selected 2 psi (13.79 kPa) as the reference pressure, and the test pressure values of 1.5, 1.75, 1.875, 2.0, 2.125, 2.25, and 2.5 psi (10.34, 12.07, 12.93, 13.79, 14.65, 15.51, and 17.93 kPa) since these pressures are a safe range of pressures for the operation of the display. Each pressure was compared against the reference ten times. 
We randomized the order in which the $P_o$ and $P$ pairs would be shown to the participant, as well as the order in which the reference and test pressure would be shown in each pair. We also showed the reference pressure against itself to measure bias on whether participants preferred choosing the first or the second pressure when unsure. 

The participants were instructed to sit at the desk, positioned in front of the curtain, and put on hearing protection headphones. Before beginning the experiments, we demonstrated the display function to the participants by inflating the display to three pressure levels and allowing them to interact with it. Each experimental trial started by inflating the display to the selected "Pressure 1". The participants were asked to touch and interact with the display for an unrestricted period of time and then release it. There was no restriction on how the participants could grasp or touch the display. Then, the display was inflated to "Pressure 2". Again, the participants were asked to touch the display and then release it. Once they interacted with both pressure levels, we asked which one felt like a higher inflation pressure. The subjects were not told the correct answers during the experiment. This procedure was then repeated until all pairs of pressures were tested ten times. Since seven different pressures were tested against the reference pressure, we had a total of 70 pairs in the study. 

After completing the interaction portion of the experiment, the participants were given a post-experiment questionnaire. The questionnaire asked about the overall experience during the study (clarity of instructions, sense of safety during the experiment) and about their previous experiences and familiarity with haptic technology, robotics, and video games. The entire experiment took approximately 35 minutes, with an optional break after the first 35 experimental pairs. 

\begin{figure}[b]
	\begin{center}
		\includegraphics[width=1.0\columnwidth]{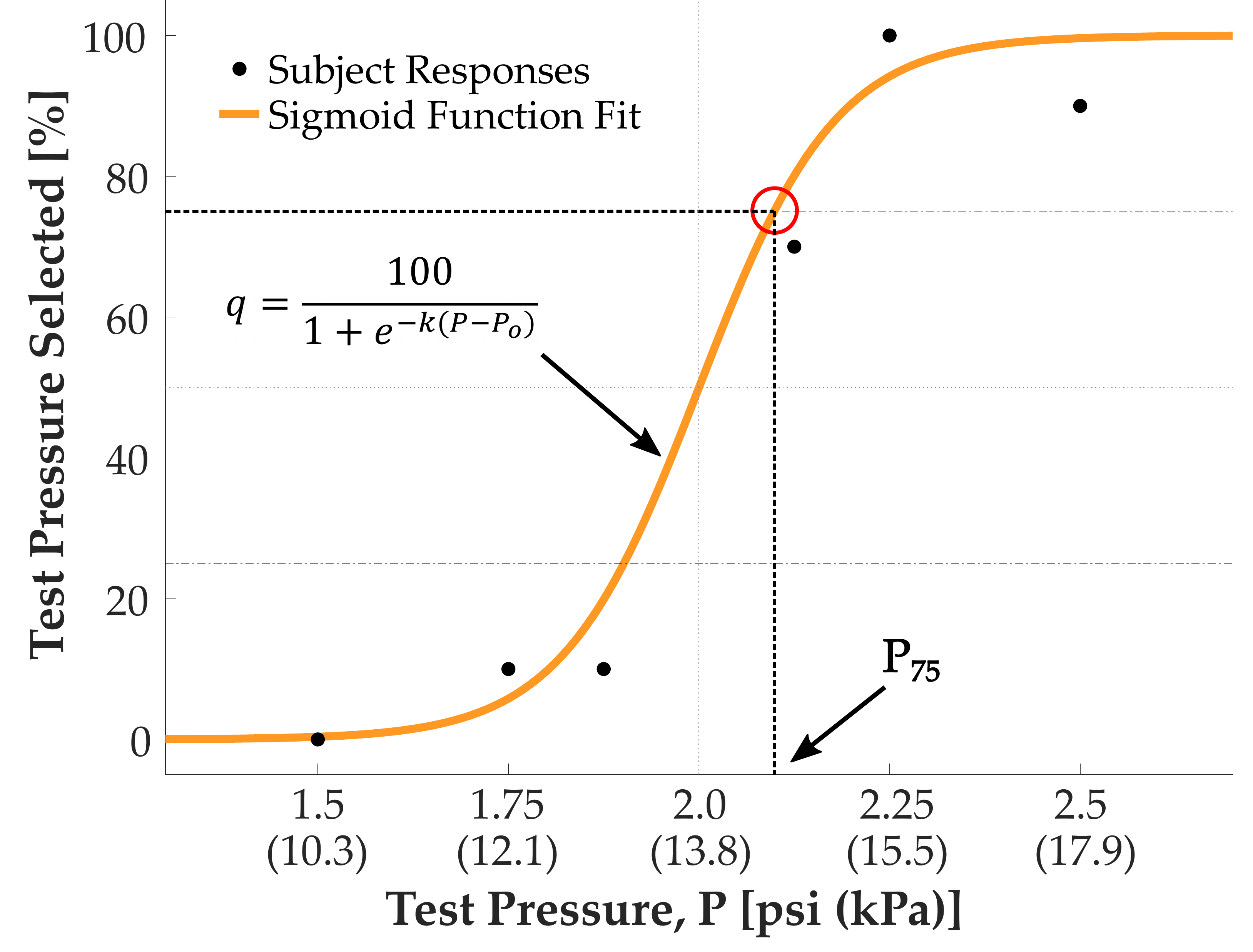}
		\caption{Raw data and sigmoid function fit for a single participant. The percentages represent the times this subject selected the test pressure, $P$, as higher. The JNDs were calculated using the sigmoid function to solve for the pressure value corresponding to the 75\% threshold, and subtracting it from the reference pressure.}
		\label{fig:single_sigmoid}
	\end{center}
\end{figure}

\subsection{Results}

A total of $10$ participants ($5$ female or nonbinary, average age $20.6$ years, age range $18-23$ years) participated in this experiment after giving informed consent. Out of the group, $9$ participants were right-handed, and $1$ was left-handed. The Purdue Institutional Review Board approved the study protocols.
Figure \ref{fig:single_sigmoid} shows a single subject's responses to the experiment. Each dot shows the percentage of time the test pressure was chosen as higher when compared against the reference pressure. The just noticeable difference (JND) was calculated by first fitting a sigmoid function to the data:
\begin{equation}\label{sigmoid}
    q = \frac{100}{1+e^{-k(P-P_o)}}
\end{equation}
where $q$ is the modeled percentage of times the user choose the test pressure ($P$) as higher, $k$ is the steepness factor for fitting a sigmoid curve, $P$ is the test pressure, and $P_0$ is the reference pressure. 
Using this fit, the JNDs are calculated by finding the pressure value corresponding to the 75\% threshold, $P_{75}$, and subtracting the reference pressure, $P_0$:
\begin{equation}\label{JND}
    JND = P_{75}-P_o = -\frac{1}{k} ln\left(\frac{100}{75}-1 \right) 
\end{equation}
Figure~\ref{fig:raw_sigmoid} shows the sigmoid function fit for each of the subjects, as well as the fit for the collection of responses from all subjects.

\subsection{Analysis}

\begin{figure}[t]
	\begin{center}
		\includegraphics[width=1.0\columnwidth]{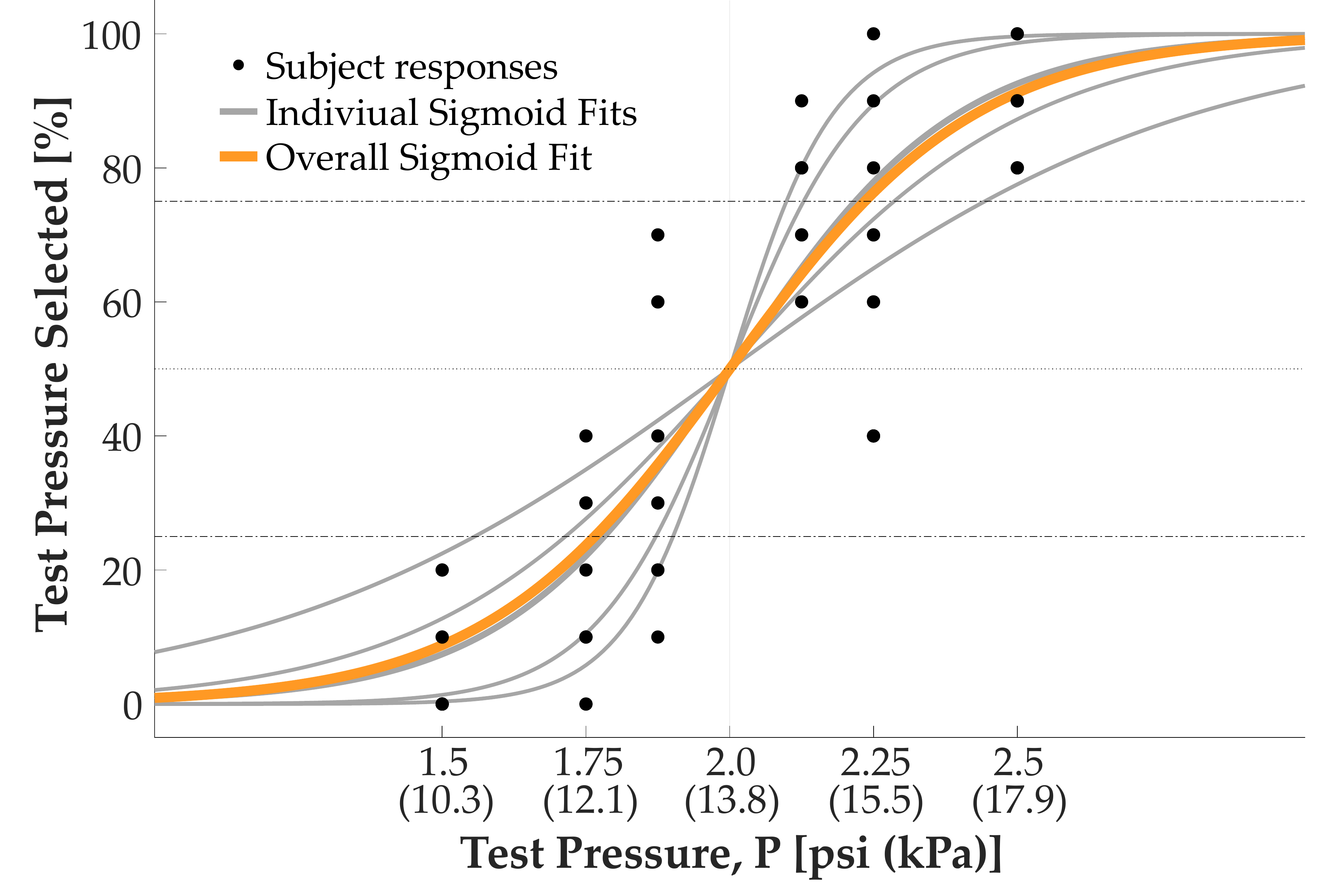}
		\caption{Sigmoid function fit for each of the subjects (grey), and the collection of responses from all subjects (orange). The dots represent percentages associated with individual subject responses. The k steepness factor for the overall sigmoid fit was 4.678, giving a JND of 0.235. The individual steepness factors ranged 2.477-11.15, with JNDs varying between 0.099 and 0.444~psi (0.68-3.06~kPa).}
		\label{fig:raw_sigmoid}
	\end{center}
	
	\vspace{-2em}
\end{figure}

The experimental results show that the $k$ steepness factor for the overall sigmoid fit (shown as the orange line in Figure~\ref{fig:raw_sigmoid}) was 4.678, with 95\% confidence bounds between 3.605 and 5.751, giving a JND of 0.235~psi (1.62~kPa). Table~\ref{tab:1} summarizes the JNDs for each of the participants. Individual JNDs ranged 0.099-0.444 psi (0.68-3.06 kPa).  The mean JND was defined as the mean of the values obtained for all participants, which was found to be 0.228 psi (1.57 kPa), with a standard deviation of 0.109 psi (0.75 kPa). The Weber fraction (WF), calculated as the ratio of the JND and the reference pressure, ranged between 4.9\% and 22.2\%, with a mean value of 11.4\%. Although there was no restriction on how the user could interact with the display, the users reported (via post-experiment questionnaire) they mainly used their fingertips. Previous studies on fingertip psychophysics tests show similar values for JNDs and WF. Frediani and Carpi \cite{frediani2020tactile} conducted psychophysical tests for a fingertip-mounted pneumatic haptic display, reporting JNDs varying in the range of 0.12-0.33 psi (0.8-2.3 kPa) for driving pressure between 0.58 and 2.90 psi (4 and 20 kPa). The WF found for this experiment was 15\%. Another study evaluating a haptic jamming display found fingertips WF to be 16\% (with a standard deviation of 7.4\%) and 14.3\% (with a standard deviation of 2.6\%) for stiffness and size perception, respectively \cite{genecov2014perception}. A different study testing stiffness perception for a rigid vibrotactile, fingertip-mounted haptic device reported WF between 17.7 and 29.9\% \cite{maereg2017wearable}. The results of this study demonstrate that our wrapped haptic display performs according to the psychometric baselines found in the literature. The JNDs and Weber fractions obtained show that the display produced detectable signals and matched previously developed rigid or soft haptic devices in performance. 

As mentioned in Section \ref{Exp Setup P}, the reference pressure was shown against itself 10 times to the subjects to measure bias on whether users had a preference for choosing ``Pressure 1'' or ``Pressure 2''. Overall, the results showed that there was no bias on their choices. The subjects chose ``Pressure 1'' as the higher pressure 45\% of the time, and ``Pressure 2'' 55\% of the time. Two subjects had a large preference for choosing ``Pressure 2'' as the highest when shown this pair of identical pressures (80\% of the time). Looking at the qualitative data, one of these subjects mentioned that they were unsure about their answers throughout the experiment, which may explain the discrepancy in their bias relative to the average bias shown by the complete pool of participants. 

The qualitative data collected from the post-experiment questionnaire shows that, besides the participant already mentioned (which had the highest JND), no other participants struggled to identify the pressures. A majority of the participants (7) mentioned that they could detect the differences and that they ``more or less agree'' or ``completely agree'' that they were sure about their answers throughout the experiment. Additionally, $9$ out of the $10$ participants said they felt safe interacting with the haptic display. It is also worth noting that the subjects with the highest correctness rates when comparing pressures mentioned they have dexterity-related hobbies or skills. For example, subject 2, who had the smallest JND and Weber fraction, mentioned that they play multiple string musical instruments. This activity requires them to vary contact pressure, which explains their high performance in the experiment. Other hand-related activities mentioned by high-performing participants include knitting, piano playing, and American Sign Language proficiency.

\begin{table}[t]
    \centering
    \caption{Experimental results for psychophysics study.}
    \label{tab:1}
    \begin{tabularx}{0.90\columnwidth}{ 
          | >{\centering\arraybackslash}X 
          | >{\centering\arraybackslash}X 
          || >{\centering\arraybackslash}X
          | >{\centering\arraybackslash}X |}
        \hline
        \textbf{Subject} &\textbf{\textit{k}}   &\textbf{JND (psi)}  &\textbf{WF (\%)}\\ 
        \hline \hline
           1        &5.048   &0.218   &10.88  \\ \hline
           2        &11.15   &0.099   &4.927   \\ \hline
           3        &3.846   &0.286   &14.28  \\ \hline
           4        &2.478   &0.443   &22.17  \\ \hline
           5        &4.989   &0.220   &11.01  \\ \hline
           6        &8.557   &0.128   &6.419   \\ \hline
           7        &2.477   &0.444   &22.18  \\ \hline
           8        &5.008   &0.219   &10.97  \\ \hline
           9        &4.574   &0.240   &12.01  \\ \hline
           10       &5.102   &0.215   &10.77  \\ \hline \hline
           \textbf{Mean}     &4.810   &0.228   &11.42  \\ \hline
           \textbf{St Dev}   &2.524   &0.109   &5.431   \\ \hline
           \textbf{Overall}  &4.678   &0.235   &11.74  \\ \hline
        \end{tabularx}
    
\end{table}

This study shows that the sensations produced by our wrapped haptic display match the psychometric measures for other haptic devices. The fingertip JNDs were in close agreement with those found in the literature. Additionally, qualitative data showed that users felt safe interacting with the display. The users were able to distinguish pressure changes without a specific task context and visual feedback. The qualitative and quantitative data show that the wrapped haptic display fulfilled the requirements outlined in Section \ref{Requirements}. Overall, we demonstrated that flexible, slim form-factor, haptic displays with soft actuation can perform as well as other haptic devices (both rigid or soft) in displaying tactile signals without encumbering normal interaction.

\section{Applying Wrapped Haptic Displays to Communicate Robot Learning} \label{VT User Study}

So far we have studied the precision with which humans can perceive the wrapped haptic display. Next, we apply this display to convey robot learning from physical interactions. In this experiment, participants kinesthetically teach a UR-10 robot arm to perform a set of cleaning tasks. We apply an existing learning algorithm to measure the robot's uncertainty \cite{menda2019ensembledagger} and then convey that uncertainty back to the human in real-time. We highlight two key differences from the experiment in Section~\ref{P User Study}: the robot arm is \textit{moving during interaction} (i.e., the wrapped haptic display is not stationary), and the haptic display \textit{now conveys a specific signal} that the human must interpret and react to during interaction.

\begin{figure}[t]
	\begin{center}
		\includegraphics[width=1.0\columnwidth]{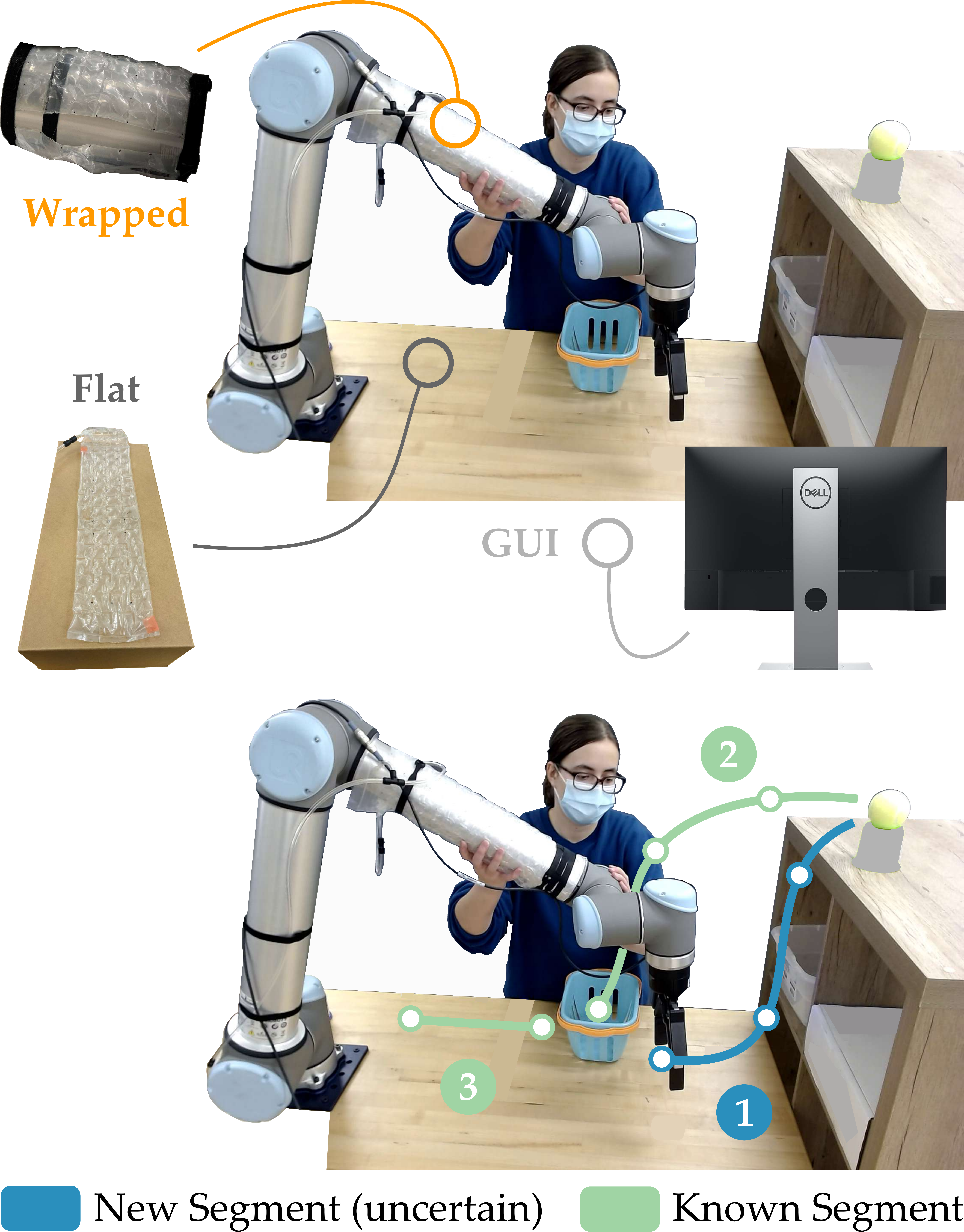}
		\caption{Participant kinesthetically teaching the robot arm the \textit{Cleaning} task. (Top) We compared our proposed approach (\textbf{Wrapped}) to two alternatives. \textbf{GUI} displayed the robot's uncertainty on a screen, while in \textbf{Flat} we placed the haptic display on table. (Bottom) We initialized the robot with data from known segments. During their first demonstration the human attempted to identify the region where the robot was uncertain (i.e., the new segment). The human then gave a second demonstration where they only guided the robot through the region(s) where they thought it was uncertain.}
		\label{fig:vt_user_study_2}
	\end{center}
	\vspace{-2.0em}
\end{figure}

\p{Independent Variables} We compared three different types of feedback (see \fig{vt_user_study_2}): 
\begin{itemize}
    \item A graphical user interface (\textbf{GUI}) that displayed the robot's uncertainty on a computer monitor.
    \item Our soft haptic display placed \textbf{Flat} on the table.
    \item Our proposed approach where we \textbf{Wrapped} the haptic display around the robot arm.
\end{itemize}
All three types of feedback showed the same information but used different modalities.
Within the \textbf{GUI} baseline we displayed uncertainty on a computer screen that was located in front of the user. Here uncertainty was shown as a percentage, where numbers close to $0\%$ meant that the robot was certain about that specific part of the task, and numbers close to $100\%$ indicated that the robot was uncertain about what it had learned. The \textbf{Flat} and \textbf{Wrapped} interfaces used the soft haptic display from Section \ref{Haptic Display and Design}. Uncertainty was linearly scaled on the haptic display from $1-3$ psi ($6.89 - 20.68$ kPa). Here $1$ psi (deflated bags) corresponded to $0\%$ uncertainty and $3$ psi (inflated bags) corresponded to $100\%$ uncertainty. The \textbf{Flat} haptic display was placed in a designated area next to the human, such that participants could periodically touch it while guiding the robot.

\p{Experimental Setup} Participants completed three different tasks with each of the three feedback conditions (i.e., nine total trials). In the \textit{Organizing} task participants were asked to guide the robot to close a drawer, pick up a ball, and then place the ball in the basket. In the \textit{Shelving} task participants kinesthetically taught the robot to close a drawer and then pull an empty container from the shelf. Finally, in the \textit{Cleaning} task participants taught the robot to pick up a ball from the top of the shelf, place it in the basket, and drag the basket to a marked location (we show the \textit{Cleaning} task in \fig{vt_user_study_2}).

Before conducting any experiments we first initialized the robot's uncertainty. We collected five expert demonstrations of each task and trained the robot with a behavior cloning approach \cite{menda2019ensembledagger}. This approach outputs the robot's uncertainty at each state (i.e., uncertainty was a function of the robot's joint position). We purposely \textit{removed} segments of the expert's demonstrations from the training set: specifically, we trained the robot without showing it how to perform either the first segment or the last segment of the task. As a result, when participants interacted with the robot, the robot was uncertain about either the start or the end of the task.

For each trial the participant provided \textit{two demonstrations}. First, the participant kinesthetically guided the robot throughout the entire task while receiving real-time feedback from \textbf{GUI}, \textbf{Flat}, or \textbf{Wrapped}. Based on this feedback, the participant attempted to identify the region of the task where the robot was uncertain (and needed additional teaching). During the second demonstration, the human \textit{only taught the segment} of the task where they believed the robot was \textit{uncertain} (i.e., the region they identified in the first demonstration). If the feedback is effective, participants should only reteach segments where the robot is confused without repeating parts of the task that the robot already knows.

\p{Participants and Procedure} We recruited ten participants
from the Virginia Tech community to take part in our study ($5$
female, average age $22.9$ years, age range $19 - 26$ years). All subjects provided
informed written consent prior to the experiment. Only one participant had prior experience physically interacting with a robot arm. Before starting the trials, we allowed participants to familiarize themselves with each task and feedback method. We used a within-subject study design: every participant interacted with all three feedback conditions. To mitigate the confounding
effect of participants improving over time, we \textit{counterbalanced}
the order of the feedback conditions (e.g., different participants start with different feedback types).

\begin{figure*}[t]
	\begin{center}
		\includegraphics[width=2.0\columnwidth]{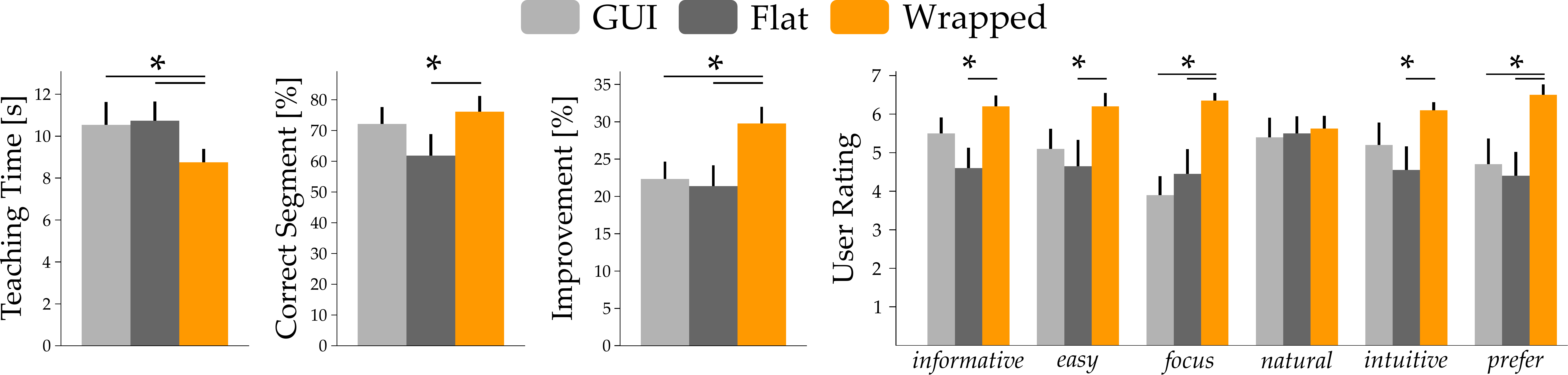}
		\vspace{-0.5em}
		\caption{Objective and subjective results when communicating robot uncertainty in real-time with \textbf{GUI}, \textbf{Flat}, and \textbf{Wrapped} feedback. Participants taught the robot three tasks; we here report the aggregated results across tasks. Error bars show standard error of the mean (SEM), and $*$ indicates statistically significant comparisons ($p < .05$). (Left) Wrapping the haptic display around the robot arm caused participants to spend less time teaching the robot, focused their teaching on regions where the robot was uncertain and improved the robot's understanding of the task after the human's demonstration. (Right) Participants thought that the wrapped display best enabled them to focus on the task, and they preferred this feedback type to the alternatives.}
		\label{fig:vt_objective}
	\end{center}
	\vspace{-2.0em}
\end{figure*}

\begin{figure}[t]
	\begin{center}
	\vspace{0.5em}
		\includegraphics[width=1.0\columnwidth]{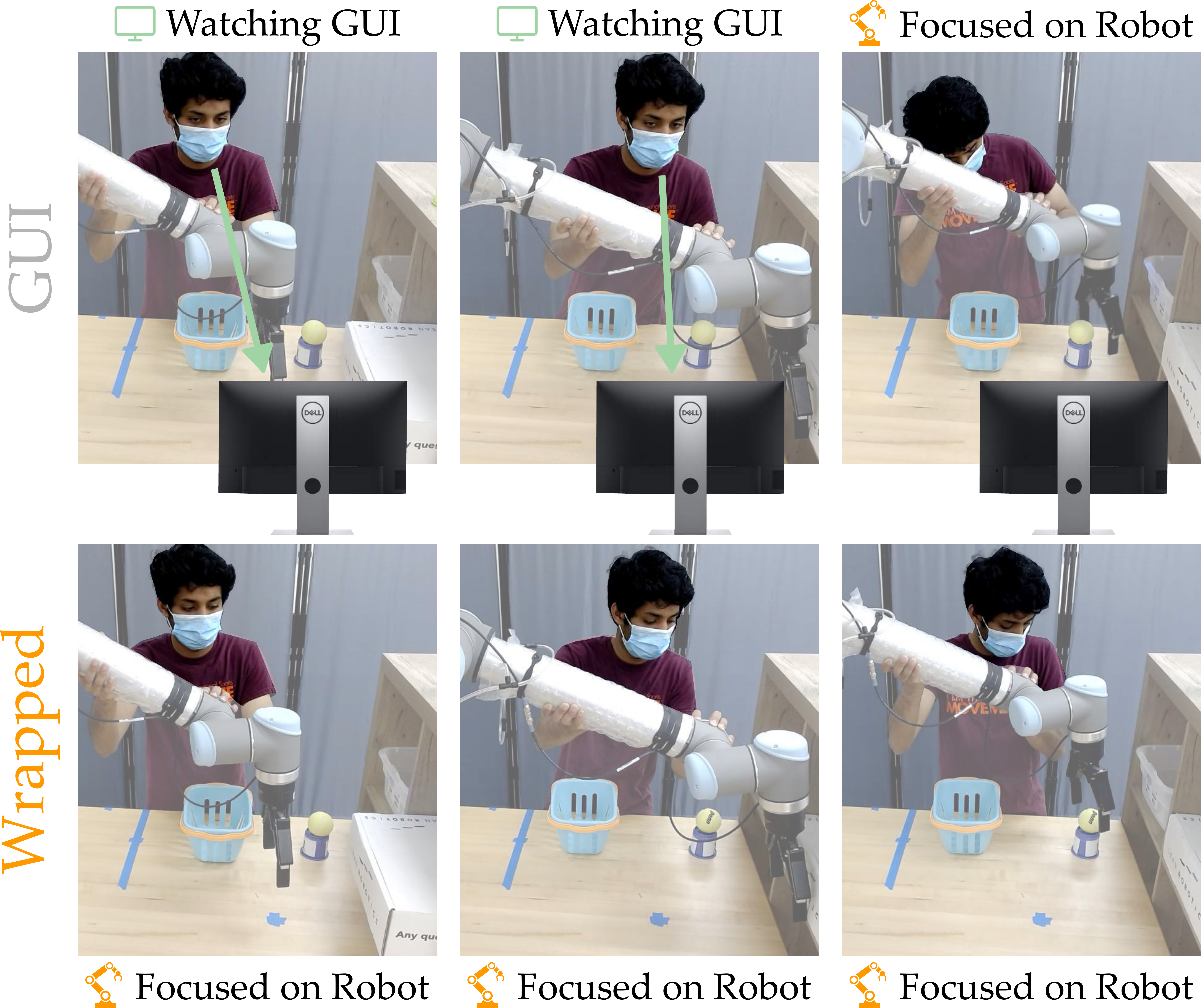}
		\caption{Participant teaching the same task under two different feedback conditions. (Top) When working with \textbf{GUI}, participants must occasionally look at the visual interface to monitor the robot's uncertainty. (Bottom) Wrapping the feedback around the robot arm enables the human to seamlessly teach the robot without having to remember to check an external interface.}
		\label{fig:vt_user_study}
	\end{center}
	\vspace{-2.0em}
\end{figure}

\p{Dependent Measures -- Objective}
Our objective measures were based on the user's \textit{second demonstration} (i.e., the demonstration where they tried to reteach the uncertain part of the task). We recorded the amount of time users spent on this second demonstration (\textit{Teaching Time}) and the percentage of this second demonstration that overlapped with the segment where the robot was actually uncertain (\textit{Correct Segment}). Offline, we also retrained the robot using the participant's second demonstration. We then measured the percentage reduction in learning uncertainty due to the user's demonstration (\textit{Improvement}).

\p{Dependent Measures -- Subjective} Participants filled out a 7-point Likert scale survey after completing all three tasks with a given method. Questions were grouped into six multi-item scales: was the user able to recognize parts they needed to repeat (\textit{informative}), did the robot’s feedback
have any effect on the user’s ability to demonstrate the task (\textit{easy}), was the user able to fully \textit{focus} on teaching the task, did the robot’s feedback seem \textit{natural} to the user, did the user find robot’s feedback \textit{intuitive} and understandable, and did the
user \textit{prefer} this current feedback method to the alternatives.

\p{Hypotheses}
We had two hypotheses for this user study:
\begin{displayquote}
    \textbf{H1.} \emph{Participants will most efficiently teach the robot with wrapped haptic displays.} 
\end{displayquote}
\begin{displayquote}
    \textbf{H2.} \emph{Participants will subjectively prefer our  wrapped haptic display over other methods.} 
\end{displayquote}

\p{Results -- Objective} We report our aggregated results in \fig{vt_objective} and show an example interaction in \fig{vt_user_study}.

We first ran a repeated measures ANOVA, and found that the robot's feedback type had a statistically significant effect on \textit{Teaching Time}, \textit{Correct Segment}, and \textit{Improvement}. Post hoc analysis revealed that participants spent less time teaching the robot with \textbf{Wrapped} than with either \textbf{GUI} or \textbf{Flat} ($p < .05$). Participants also better focused their teaching on the region where the robot was actually uncertain: \textbf{Wrapped} resulted in a higher \textit{Correct Segment} than \textbf{Flat} ($p < .05$). However, here the differences between \textbf{Wrapped} and \textbf{GUI} were not statistically significant ($p=.287$).

Recall that \textit{Improvement} captures how much more confident the robot is about the task after the participant's demonstration. This metric is especially important: we want to enable humans to teach robots efficiently, and \textit{Improvement} quantifies how much the robot learned from the human's teaching. We found that the robot's confidence improved the most in the \textbf{Wrapped} condition as compared to either \textbf{GUI} or \textbf{Flat} ($p < .05$). Overall, these results support \textbf{H1}: when users get real-time feedback from a haptic display wrapped around the robot arm, they provide shorter duration kinesthetic demonstrations that more precisely hone in on the robot's uncertainty and efficiently correct the robot.

To better explain why \textbf{Wrapped} outperformed \textbf{GUI}, we include an example interaction in \fig{vt_user_study}. Notice that --- when the feedback was not located directly on the robot arm --- participants had to periodically turn their attention away from the task in order to check the robot's uncertainty. For \textbf{Flat}, this required taking a hand away from the robot and feeling the haptic display on the table; for \textbf{GUI}, participants had to look up and check the computer monitor. The key difference with \textbf{Wrapped} is that this haptic display is located at the point of interaction, and thus participants could experience feedback while still remaining focused on the task and their physical demonstration.

We were initially surprised that --- although users with \textbf{Wrapped} and \textbf{GUI} scored similarly for \textit{Correct Segment} --- the results for \textit{Improvement} were significantly different. However, we believe the explanation for this lies in the quality of the participants' demonstrations. Returning to \fig{vt_user_study}, we recognize that with \textbf{GUI} participants often had to pause and check the uncertainty, breaking up their demonstration (and causing the demonstration to include multiple stops). Our subjective results support this explanation: as we will show, participants reported that they were more distracted with \textbf{GUI} than with \textbf{Wrapped} feedback.

\p{Results -- Subjective} Figure \ref{fig:vt_objective} depicts the results from our Likert scale survey. After confirming that our six scales were reliable (using Cronbach's alpha), we grouped these scales into combined scores and ran a one-way repeated measures ANOVA on each resulting score.

Participants perceived each of the feedback methods as similarly natural. But post hoc analysis showed that participants thought that \textbf{Wrapped} was more informative, easier to interact with, less distracting, and more intuitive than either one or both of the alternatives ($p < .05$). Participants also indicated that they preferred \textbf{Wrapped} over \textbf{GUI} and \textbf{Flat}. When explaining this preference, one participant said, \textit{``I definitely prefer \textbf{Wrapped} over other methods. I was able to clearly focus and the other methods were distracting.''}. Our subjective results support \textbf{H2}, and indicate that users perceived wrapped haptic displays as preferable when compared to alternatives like visual interfaces.

\p{Limitations} This experiment is a first step towards wrapped haptic displays that communicate physical robot learning. During the user study we purposely caused the robot to be uncertain about either the first segment (the start of the task) or the last segment (the end of the task). Anecdotal evidence suggests that some participants expected the robot to always know the start of the task: \textit{``How can a robot be confused at the beginning?''} We recognize that this assumption may be a confounding factor in our results (although we did randomize the unknown segment across tasks and users).

\section{Conclusion}

In this paper we have presented a soft wrapped haptic display capable of communicating information about a robot's internal state during physical interaction. We manufactured the pneumatic device using soft, flexible pouches that render haptic signals through pressure, and then wrapped this haptic display around rigid robot arms. Our results suggest that humans can accurately distinguish between different pressures rendered by the wrapped haptic display, and that this approach provides more informative feedback about robot learning than current alternatives. Future work will expand the complexity of the signals rendered by the soft display: we will alter the pouch shapes and distributions, and exploit these enhanced capabilities of our soft array to provide more nuanced and complex feedback.


\bibliographystyle{IEEEtran}
\bibliography{references}
\end{document}